# Edge and Corner Detection for Unorganized 3D Point Clouds with Application to Robotic Welding

Syeda Mariam Ahmed[1], Yan Zhi Tan[2], Chee Meng Chew[1], Abdullah Al Mamun[2], Fook Seng Wong [3]

***Abstract*—In this paper, we propose novel edge and corner detection algorithms for unorganized point clouds. Our edge detection method evaluates symmetry in a local neighborhood and uses an adaptive density based threshold to differentiate 3D edge points. We extend this algorithm to propose a novel corner detector that clusters curvature vectors and uses their geometrical statistics to classify a point as corner. We perform rigorous evaluation of the algorithms on RGB-D semantic segmentation and 3D washer models from the ShapeNet dataset and report higher precision and recall scores. Finally, we also demonstrate how our edge and corner detectors can be used as a novel approach towards automatic weld seam detection for robotic welding. We propose to generate weld seams directly from a point cloud as opposed to using 3D models for offline planning of welding paths. For this application, we show a comparison between Harris 3D and our proposed approach on a panel workpiece.**

## I. Introduction

Visual features like edge, corner and texture are widely used in computer vision and robotic perception. They can be detected through low-level cues and then subsequently used for higher level perception tasks. A rapidly growing application is robotic welding which depends on extraction of corners and edges to generate welding paths for the robot. An example of a welding workpiece, known as panel workpiece, is shown in Fig. 1. Existing approaches for such welding tasks are based on offline simulation and planning using 3D CAD models. With the growing demand for automation, alternative solutions are required that are more open to adaptation to previously unseen and complicated welding tasks.

Though a 3D point cloud is more informative as compared to a 2D image, extracting edge and corner features in 3D is much more challenging. In case of edge detection for unorganized 3D point clouds, conventional methods perform surface reconstruction to form a mesh [1, 8, 16, 17] or build a graph [3] to analyze the neighborhood of each point through principal component analysis (PCA). However, reconstruction tends to wash out sharp edges and intricate features while graph based methods are computationally expensive [7]. The most recent approach, applicable to unorganized point clouds, is based on discriminative learning [2]. The authors train a random forest based binary classifier on a set of features that learn to classify points into edge versus non-edge points. One drawback of their method is poor performance on unseen data.

Real-time state-of-the-art methods for 3D edge detection are much more efficient for organized point clouds [5, 6]. This is due to the fast nearest neighbor search that can be performed in an organized structure and makes them suitable for real-time applications. However, their fast speed is a result of the nearest neighbor search in a regular 3D grid and thus their performance is restricted to organized point clouds.

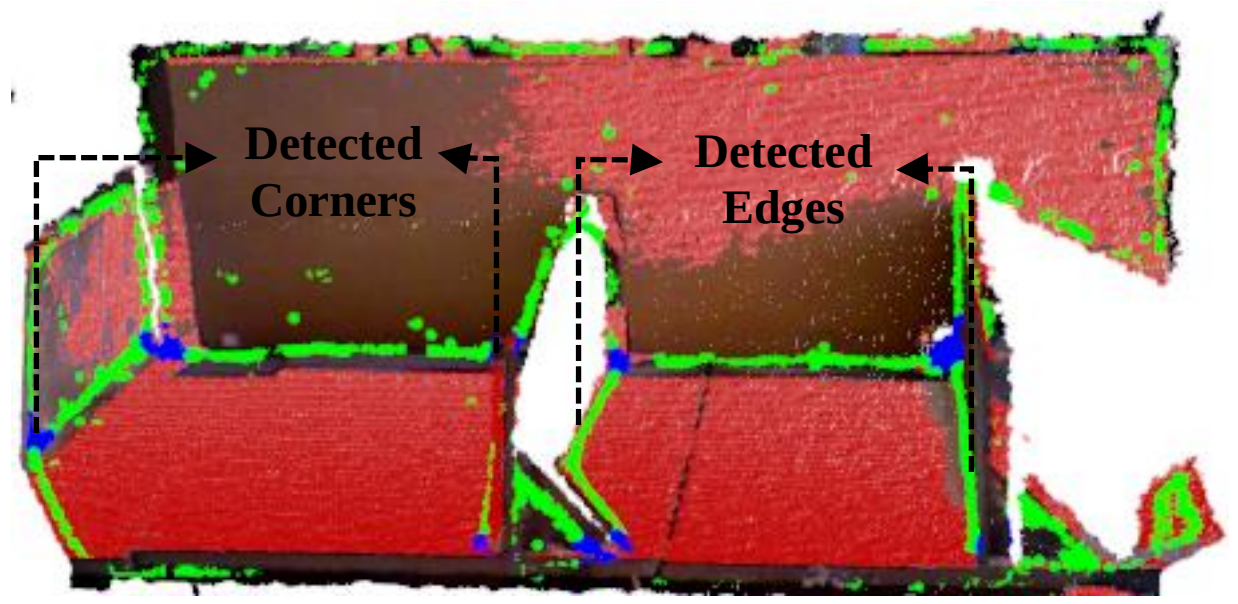

Fig. 1. Demonstration of our edge and corner detectors on 3D pointcloud of panel workpiece.

Similarly corner detection methods in 3D are often an extension of 2D techniques [9] and require an organized structure or a depth image for detection [12]. Additionally, various keypoint detection algorithms exist in 3D, however,they aim to find stable and repeatable points that are not directly on the edge [10, 11]. Thus it is not possible to use such methods for accurate corner detection.

This paper has two main contributions: the first is a new edge detection algorithm; second is a novel corner detector. The edge detection method evaluates the level of symmetry in a local region of a 3D point. A main advantage of this approach is that it is independent of surface normals due to which the detected edges show a higher accuracy. We extend the edge detection algorithm to differentiate between line-type or sharp features and propose a novel corner detector. Our approach is most similar to the Gauss map clustering method, proposed by Weber [3]. They observe that sharp features separate two or more surfaces and in presence of a discontinuity, the points of a sphere will build more than one distinct cluster. Using a similar intuition, our corner detection approach is also based on clustering of curvature vectors. However the key differences are: 1) we only evaluate edge points in the point cloud, thus reducing the number of points to search for corners. 2) We compare features extracted from the size and average curvature vector associated with each cluster to detect a corner, as opposed to evaluating the number of clusters formed. As a result, our proposed method can detect accurate corners.

## II. 3D Edge Detection

The proposed edge detection method is based on the mean shift algorithm [18] which is a non-parametric iterative

Syeda Mariam Ahmed[1] and Chee Meng Chew[1] are with the Department of Mechanical Engineering, National University of Singapore (email: mpesyed@nus.edu.sg ). Yan Zhi Tan[2] and Abdullah Al Mamun[2] are with the Department of Electrical Engineering, National University of Singapore. Fook Seng Wong[3] is with Keppel Technology and Innovation, Singapore (e-mail: fookseng.wong@komtech.com.sg).

technique to locate the maximum density/modes of a function in feature space. Our algorithm for edge detection in 3D point cloud uses the same intuition, but instead of finding the local maxima, we evaluate the degree of shift in centroid from its initial position to classify points as edge or non-edge.

More formally, we define the approach as follows. Given a query point $p_i$, we determine its k-nearest neighbors. For an unorganized point cloud, this is achieved through a k-dimensional (K-d) tree [13]. For an organized point cloud, the nearest neighbors are all points in a square of size $s$, while $p_i$ is the centroid. These neighboring points of $p_i$ are given as $V_i = \{n_1, n_2, \dots. n_k\}$. Initially we assume that the centroid of $V_i$ is the query point itself, while a new centroid $C_i$ is computed by taking the mean of the neighboring points as follows:

$$C_i = \frac{1}{|V_i|}\sum_{j=1}^{k} n_j \quad (1)$$

To cater variation in density, we compute the resolution $Z_i(V_i)$, as defined by (2), of the neighboring points. This is achieved by determining the distance of the nearest neighbor of $p_i$ among all the $k$ neighbors.

$$Z_i(V_i) = \min_{n \in V_i} \|p_i - n_i\| \quad (2)$$

$$\|C_i - p_i\| > \lambda \cdot Z_i(V_i) \quad (3)$$

Evaluation of $Z_i$ ensures scale invariance as the local density of points is considered for each point individually. Finally, $Z_i$ is multiplied by a fixed parameter $\lambda$ which serves as the classification threshold. If the distance between the new centroid $C_i$ and the query point $p_i$ is greater than this threshold times $Z_i(V_i)$ as defined by (3), the point is classified as an edge.

Fig. 2 illustrates the concept behind the proposed approach on a partial view of the *Stanford bunny* model. It can be seen that the neighboring points $V_i$ (shown in green) of the query point $p_i$ (shown in red), do not have a symmetrical circular shape, but rather ends abruptly at the edge. This results in a larger shift in the position of the centroid $C_i$ (shown in blue). As opposed to this, a point on a smoother surface will have a shift in centroid position that is comparatively small. This fundamental idea of symmetry among neighboring points can enable quick classification of 3D data.

The only variable parameters in the algorithm are $k$ and $\lambda$, that determine the number of nearest neighbors and the classification threshold, both of which can be easily tuned for any point cloud. For data with lower signal to noise ratio, a higher value of $k$ and $\lambda$ will be required and vice versa.

## III. Corner Detection

Efficient corner detection algorithms are the basis for many vision based applications. A 3D corner can be defined at the intersection of two or more edges. Theoretically, there should not be ambiguity in the location of the corner as it is spatially constrained, however noisy point clouds can make it difficult to accurately localize the corner point. To this end we propose a novel corner detector for 3D point clouds.

The corner detection algorithm is mathematically formulated as follows. Given a point cloud $\mathcal{P}$, we detect edges using the method described in Section II. As a result the data points are classified as edge $E$, and non edge points $\mathcal{N}$ such that $\mathcal{P} = (E, \mathcal{N})$. For all points in $E$, we compute their curvature defined by the Eigenvector $e$ of the smallest Eigenvalue, $\lambda_1$, determined from the covariance tensor $\Sigma_E$ of the neighboring points in $E$. The covariance tensor used is a weighted linear combination of neighboring points proposed by [19]. The covariance tensor is defined as follows:

$$\Sigma_E = \frac{1}{\sum(R - d_i)}\sum_{i=1}^{n}(R - d_i)(p_i - \bar{p})(p_i - \bar{p})^T \;\; \forall\, p \in E \quad (4)$$

where $R$ is a predefined radius, $d_i = \|p_i - p\|_2$ and $\bar{p}$ is the centroid of the neighboring points $p \in E$. The advantage of this tensor is that it assigns smaller weights to distant points so as to increase repeatability in presence of clutter.

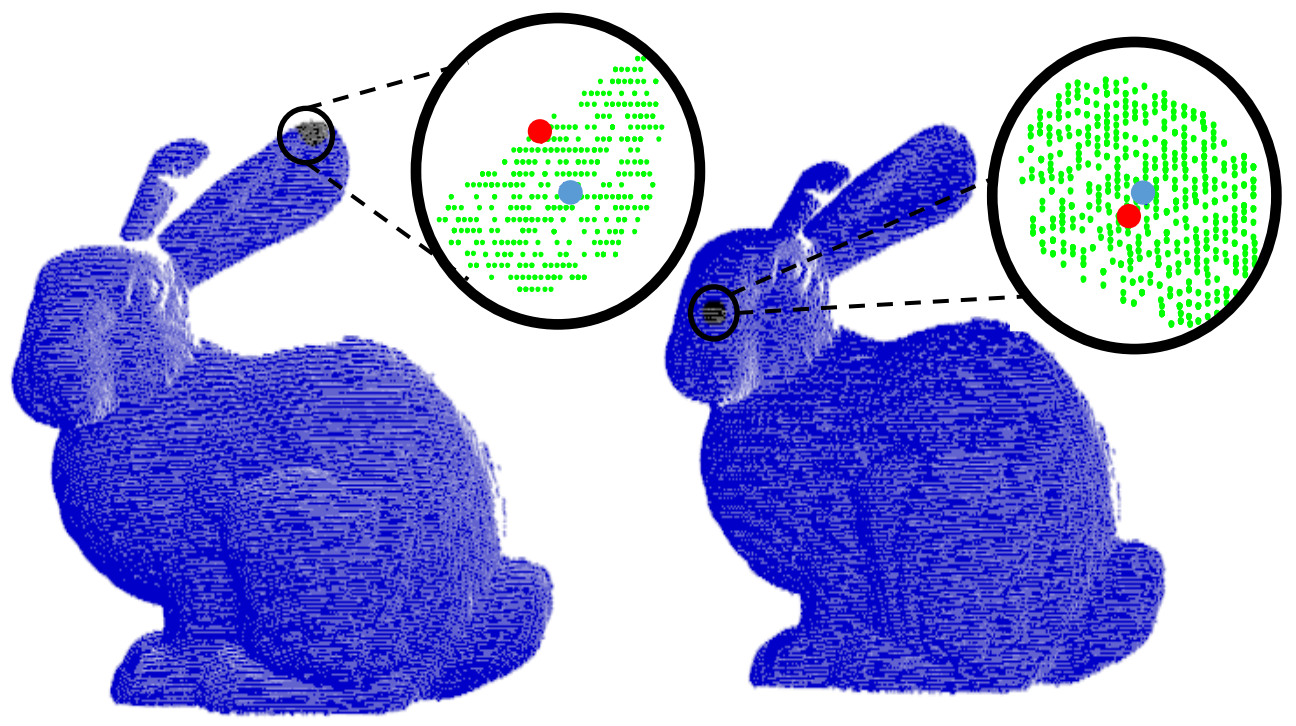

Fig. 2. Demonstration of shift in centroid for edge and non-edge points for a partial view of the Stanford bunny model. The query point $p_i$ is shown in red, the neighboring points $V_i$ are shown in green and the new centroid $C_i$ is shown in blue.

Next, for each point $p_i \in E$, we find its $k$ nearest neighbors from $E$ defined as follows:

$$\mathcal{K} = \{u_1, u_2, \dots. u_k\} \;\; u \in E, |\mathcal{K}| = k \quad (5)$$

Since corners in 3D may exist at the intersection of numerous edges, we assume to be dealing only with objects where the corners lie at intersection of either two or three edges. Thus it is important to distinguish between the type of corner, so as to define the required number of clusters. This check is performed by evaluating the maximum variation in Euclidean distance within each of the *x*, *y* and *z* directions as follows: .

$$N(\mathcal{K}_i)\begin{cases} \left|\max\limits_{1<i<k} u_i(x) - \min\limits_{1<i<k} u_i(x)\right| > \rho \\ \left|\max\limits_{1<i<k} u_i(y) - \min\limits_{1<i<k} u_i(y)\right| > \rho \\ \left|\max\limits_{1<i<k} u_i(z) - \min\limits_{1<i<k} u_i(z)\right| > \rho \end{cases} \quad (6)$$

where $N(\mathcal{K}_i)$ is the number of clusters for the group of points that belong to $\mathcal{K}_i$ while $u_i(x)$, $u_i(y)$ and $u_i(z)$ are the *x*, *y* and *z* values of each point $u_i$. $\rho$ is a fixed parameter that determines the minimum variance along each direction to indicate the existence of a corner at that particular location. If the variation in all three directions is above $\rho$, the corner is predicted to be located at the intersection of three edges. Similarly, if the variation is significant in two directions, the given corner will most likely exist at the intersection of two edges. Since our corner detector is based on a clustering approach, this check determines the number of clusters $N$, for splitting features.

Our features associated with each cluster are 1) size of the cluster and 2) mean curvature vector of the cluster. These are shown in Fig. 3a) and 3b) where a corner of a washer (from the ShapeNet dataset [14]) lies at the intersection of three edges. In this case, since the corner belongs to a regular shaped cuboid, the mean curvature vectors $\mu_i$ associated with each cluster should be approximately orthogonal to each other. Thus, an angular variation between curvature vectors of the clusters in this range will indicate existence of such corners.

Similarly, the size of clusters is an indication of how far the query point is from the actual corner. Considering the cube in Fig. 3a), a corner point will consist of k-nearest neighbors that have an equally distributed number of points with curvature vectors in three unique directions. The equal sized clusters are shown in red, yellow and purple. When the size of the three clusters is relatively equal, we consider the query point to be close to the actual corner. However, as shown in Fig. 3b), as the query point moves away from the real corner, its k-nearest neighbors will have points that consist of a bias towards a particular curvature direction. Thus, one of the clusters will be comparatively larger than the other two. This indicates that the query point is far away from the actual position of the corner. Thus, these properties can be used to distinguish well localized corners from edge points.

Mathematically, we formulate these concepts as follows. Given the number of clusters from our initial condition, the neighboring points $u_i \epsilon E$ are partitioned into $N$ sets $S = \{S_1,.S_n\}$, $n = 2,3$ using k-means algorithm, based on the associated curvature vectors $e$. K-means algorithm will minimize the within-cluster variance among the curvature vectors which is defined as:

$$\min_{S} \sum_{i=1}^{k} \sum_{e \in S_i} \|e - \mu_i\|^2 \quad (7)$$

where $\mu_i$ is the mean of curvature vectors for its respective cluster $S_i$. The geometrical properties associated with each cluster are their mean curvature vectors $\mu_i$ and the size of each cluster $S_i$. We use these properties to extract two prominent features a) angular variation in $\mu_i$ defined as $\phi_{ij}$ in (9) and b) difference in size of the clusters denoted as $|S_i - S_j|$. We evaluate the angular variation in curvature between two clusters as follows:

$$\phi_{ij} = atan2\left(\|\overrightarrow{\mu_i} \times \overrightarrow{\mu_j}\|, \overrightarrow{\mu_i} \cdot \overrightarrow{\mu_j}\right) \;\; i,j = \{1,2,3\}, i \neq j \quad (8)$$

where $\overrightarrow{\mu_i}$ and $\overrightarrow{\mu_j}$ are mean curvature vectors for $S_i$ and $S_j$ respectively. Thus the feature vector $\phi_{ij}$ represents the angular variation in curvature for all $N$ clusters. The second property is the difference in size of the sets given as $|S_i - S_j|$ where $i,j = \{1,2,3\}, i \neq j$. Using these features, we classify the query point $p_i$ as a corner if it satisfies the following conditions:

$$C(p_i) \begin{cases} 1, \begin{cases} |S_i - S_j| < \varepsilon, & i,j = \{1,2,3\}, i \neq j \\ \theta_1 < \phi_{ij} < \theta_2, & \theta_1, \theta_2 \in [0, \pi] \end{cases} \\ 0, \quad \text{otherwise} \end{cases} \quad (9)$$

where $C(p_i) = 1$ indicates a corner, $\varepsilon$ is a slack variable while $\theta_1, \theta_2$ define the allowable range of angular variation between the Eigenvectors. The introduction of the slack variable $\varepsilon$ is meant to add robustness in case of noisy point clouds and allows the flexibility to classify a group of points as corners and finally use their centroid as the true corner.

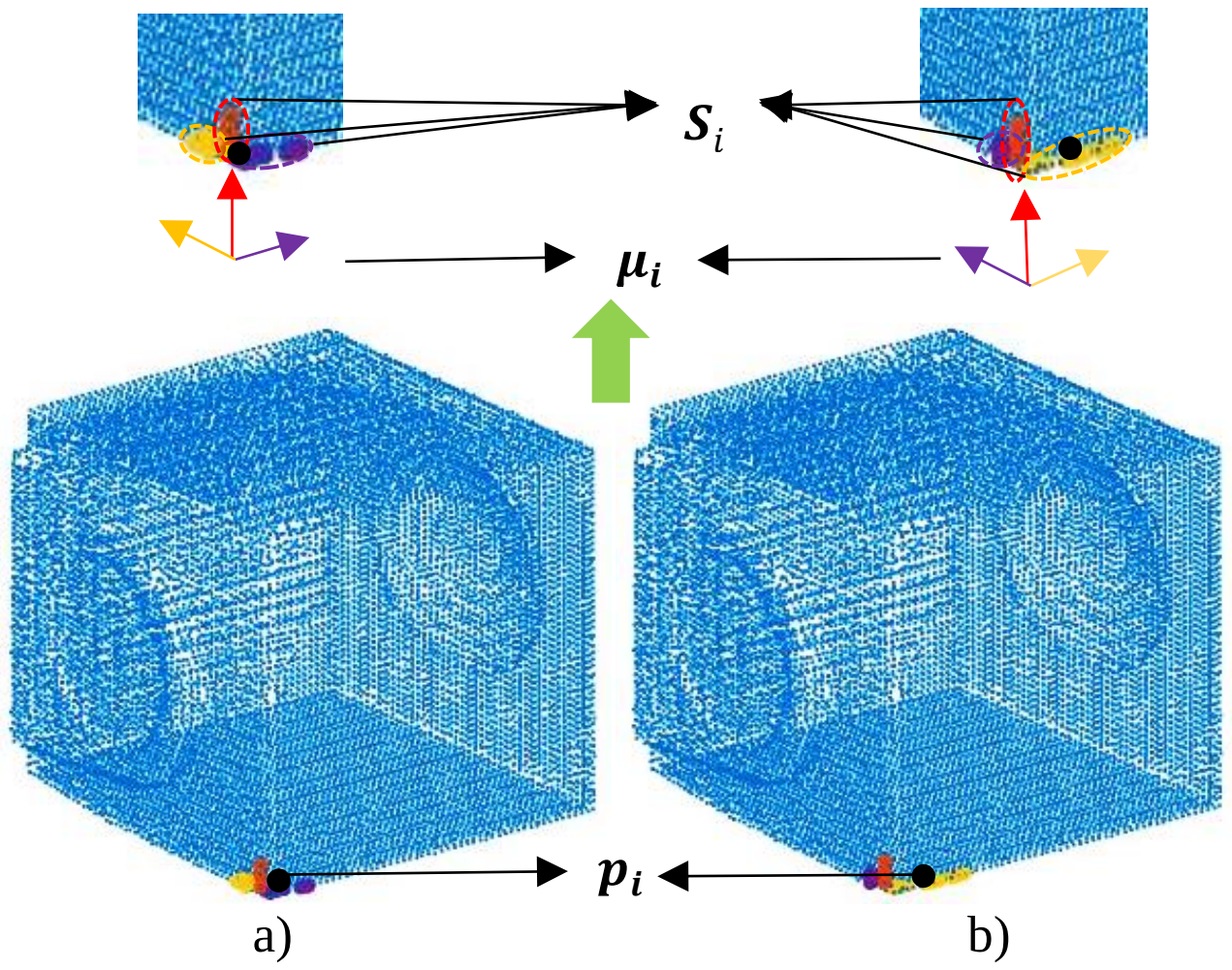

Fig. 3. Demonstration of a corner that lies at the intersection of three edges, a) accurate localization of corner as the size of clusters $S_i$ is similar and b) poor localization of corner as the yellow cluster is significantly larger.

## IV. Experimental Results

In the first section of our results, we evaluate the proposed algorithm for edge detection against state-of-the-art edge detection algorithms for organized and unorganized point clouds. We demonstrate our results on the RGB-D semantic segmentation dataset [20] for comparison. In the next section we evaluate the repeatability and accuracy of the corner detector on 3D models of washers from the ShapeNet dataset [14]. Finally, we show how the algorithms proposed above can be used to automate welding of a panel workpiece. All experiments described in the following sections are run on an Intel i7-4600M CPU with 2.9 GHz and 8GB RAM. No multi-threading or any other parallelism such as OpenMP or GPU was used in our implementation.

### *A. Evaluation of Edge Detection*

The 3D edge detection algorithm was evaluated using the RGB-D semantic segmentation dataset [20] that is acquired using the Microsoft Kinect device. The dataset provides 3D meshes and yaml files for 16 different scenes that includes 5 categories of common grocery products such as packets of biscuits, juice bottles, coffee cans and boxes of salt. We use the 3D meshes to generate ground truth using Meshlab [15] that directly allows edge selection for the dataset. In addition, we use the yaml files to extract organized point clouds for comparative evaluation.

We use three edge detection algorithms for comparison [5-7]. We will refer to the first two algorithms as D+HC and D+SC, respectively, throughout the rest of the paper. Both of these algorithms are available online as part of the PCL [4]. Their limitation is that they can only be implemented on organized point clouds. The third method is based on evaluating Eigenvalues of the covariance matrix that is defined by each point's local neighborhood [7]. We will refer to this algorithm as EV. Unlike the first two methods, this alg-orithm can be implemented on both organized and unorganized point clouds. Finally, we will refer to our proposed edge detection algorithm as MS, Mean Shift based edge detector.

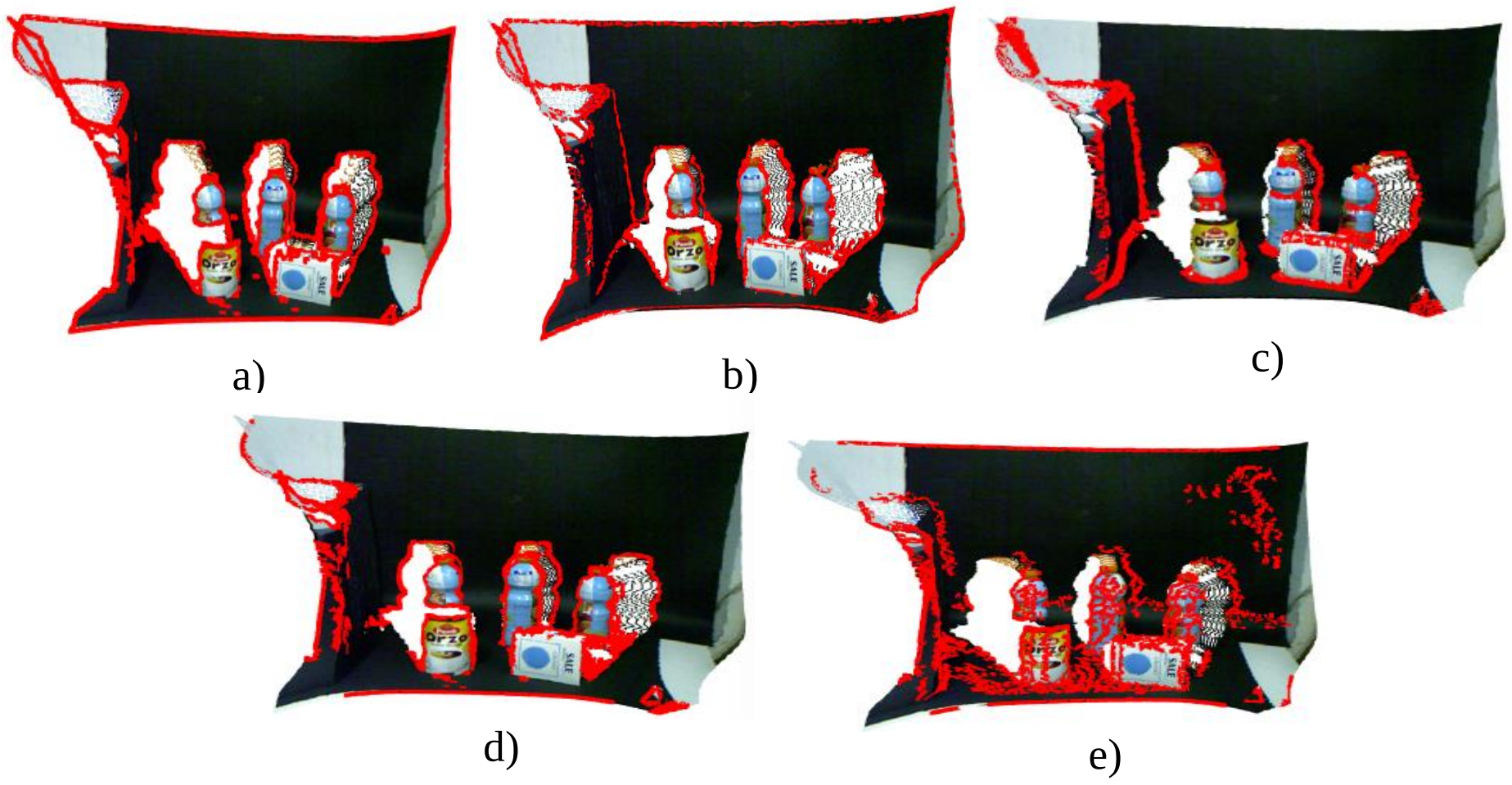

Fig. 5. Comparison of all four edge detection algorithms on a scene from the RGB-D semantic segmentation dataset where a) ground truth extracted from the mesh of the scene, b) results of edge detection from the proposed approach, c) results of edge detection using EV [7], d) results of edge detection using D+HC [6] and e) results of edge detection using D+SC [5].

Quantitave analysis of the results is performed using the precision and recall estimates. We conduct experiments with a nearest neighborhood of k = {10, 50, 100, and 150} (this parameter is consistent for all algorithms). For the particular dataset, it was observed that best results were obtained for k = 100 for all algorithms. For D+HC and D+SC, the main parameter is the depth discontinuity, *d*, which is evaluated over the range of $0.0002 < d < 0.5$ and $0.0001 < d < 1.5$ for each method respectively. In addition, D+SC also requires parameter tuning for Gaussian filter, detectable edge angle and line smoothing parameters. We experimented with Gaussian kernel sizes of {3x3, 7x7, 10x10, 20x20 and 40x40}. Similarly we varied edge angle for a range of $\theta \in \{10°, 20°, 25°, 30°, \text{and } 40°\}$ while after experimenting with a few different values, the minimum and maximum line width was kept at 10 and 30 respectively.

From our experiments, we obtained the best results at $d = 0.0002$ and 0.0001 for D+HC and D+SC respectively. Additionally, a Gaussian kernel size of 40x40, $\theta = 10°$ and maximum and minimum line width at 10 and 30 were found to be optimal parameters for the given dataset. For EV, the only variable parameter is the factor of surface variation $\sigma_k(p)$ that is used to distinguish between edge versus non- edge points. This parameter was varied between $0 < \sigma_k(p) < 27$ and highest precision was achieved at $\sigma_k(p) = 27$. Finally, our proposed algorithm MS was evaluated for the threshold parameter varying between $0 < \lambda < 8$ and optimal precision was achieved at $\lambda = 8$.

Precision and recall results for the parameters described above, are shown in Fig. 4. The graph clearly shows that MS has a consistently higher precision and recall as compared to all other algorithms. D+HC shows high precision in a very small range after which it quickly drops. Similarly EV and D+SC demonstrate even lower precision values that indicate a low level of TPs for generally majority of their parameter range. These results are validated by a visual analysis of detected edges, shown in Fig. 5. It can be seen that EV and D+HC tend to miss out boundary edges, which results in lower precision. Additionally, D+SC generates noisier edges that further reduces TP and increases FP. As opposed to all three algorithms, results from the proposed algorithm are more precise and most similar to the ground truth edges.

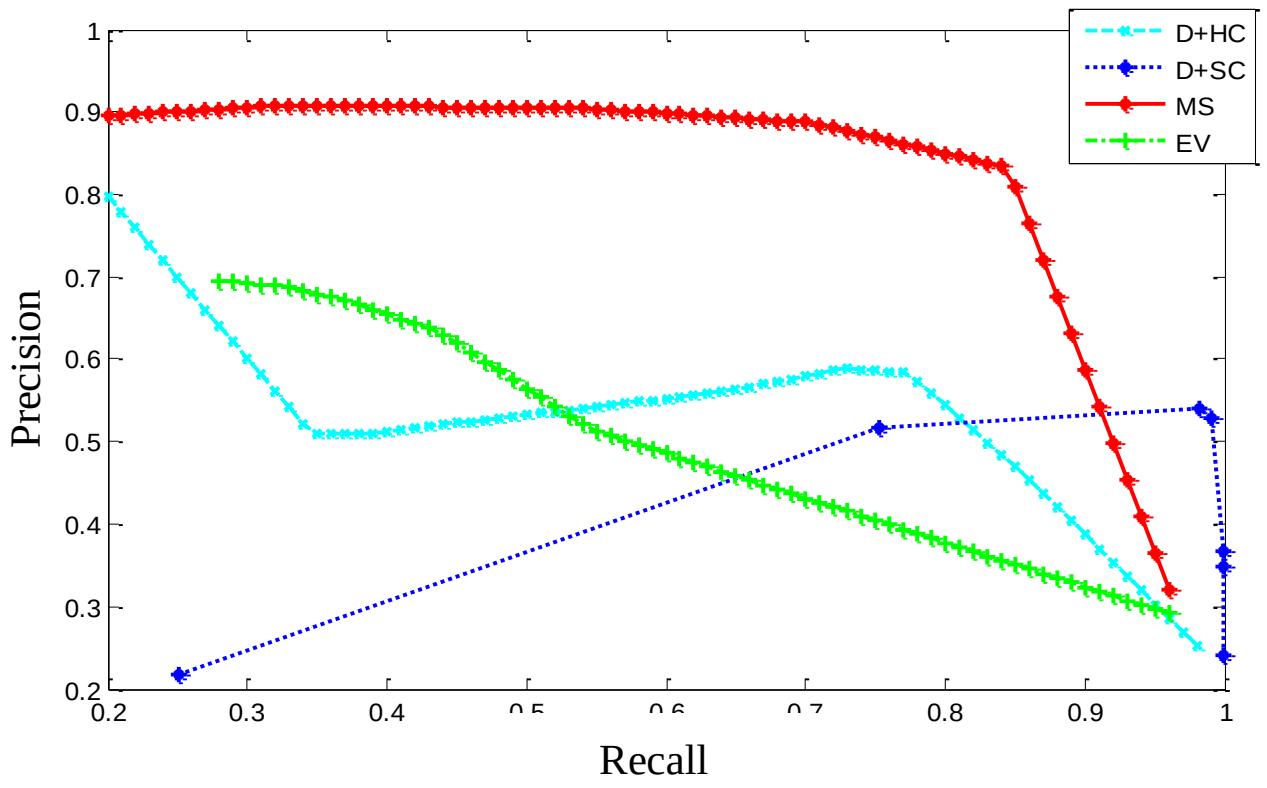

Fig. 4. Precision recall curve for proposed method, EV, D+HC and D+SC, with the fixed parameter of k = 100.

D+HC and D+SC still demonstrate faster computation times of an average time of 217 ms and 700 ms for one point cloud of size 640x480x3. After downsampling the cloud with a voxel grid leaf size of 0.005x0.005x0.005 and employing Kd tree search with k=100, for unorganized point clouds of size 307,200 points, EV and MS took an average of 4303 ms and 3515 ms respectively. We believe that the code can be optimized to reach much faster computations in future.

### *B. Evaluation of Corner Detector*

In this section we evaluate the accuracy of the corner detector. For testing we use ten 3D models from the washer category of the ShapeNet dataset [14]. The corners in these 3D models are manually labeled using Meshlab.

The first experiment analyzes the accuracy of corner detection by varying $\varepsilon$ for various $\mathcal{K}$ nearest neighbors from the edge cloud $E$. For this experiment, the fixed parameters

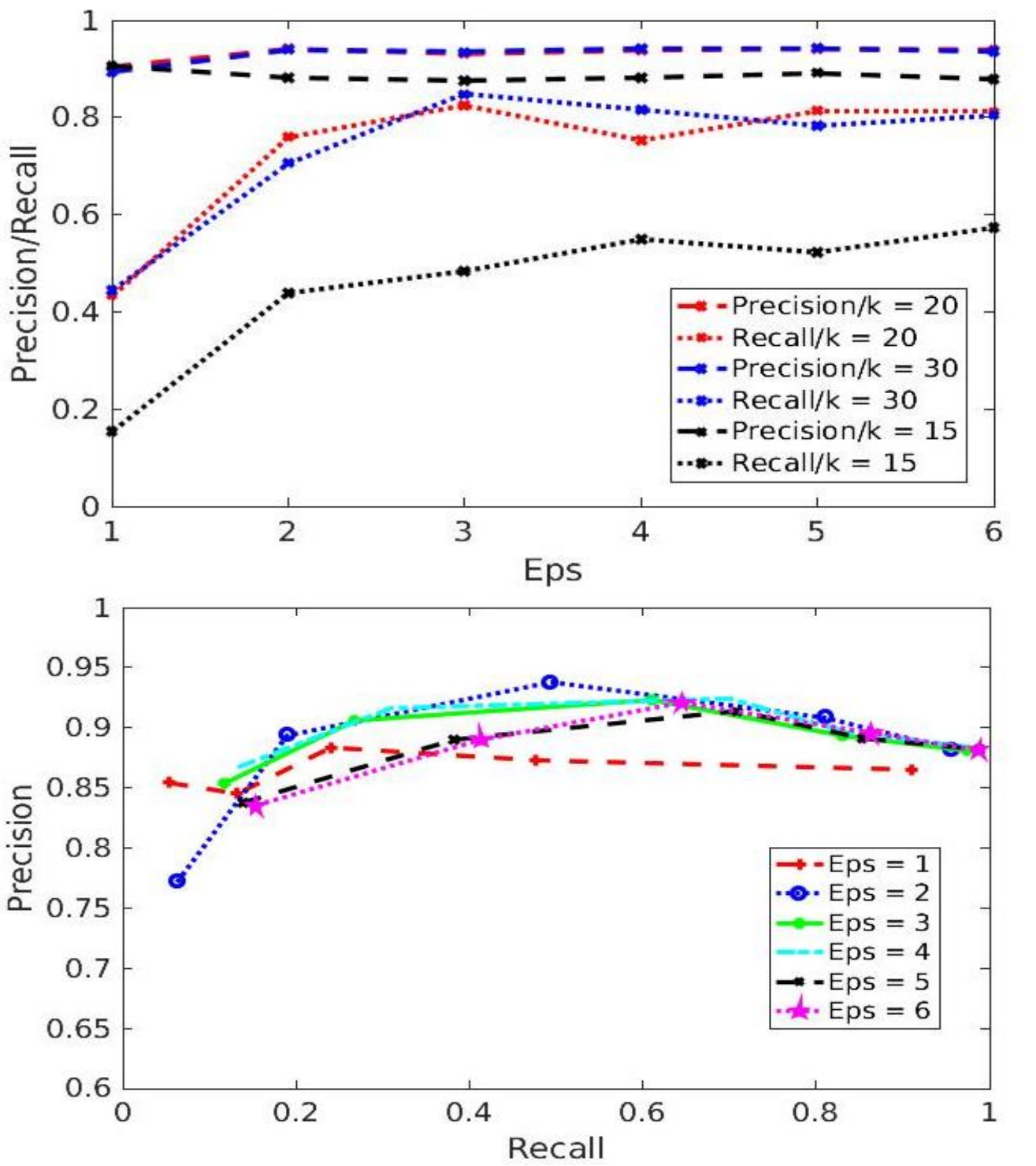

Fig. 6 a) Precision recall curve versus $\varepsilon$ for different values of $\mathcal{K}$ nearest neighbours and b) precision recall curve for variable $\varepsilon = \{1\text{-}6\}$, $30^\circ < \theta_1 < 80^\circ$ and $110^\circ < \theta_2 < 140^\circ$.

were $\rho = 0.005$, $\theta_1 = 60^\circ$ and $\theta_2 = 130^\circ$ while the variable parameters were $1 < \varepsilon < 6$ and $\mathcal{K} = \{15, 20, 30\}$. As seen from the graph in Fig. 6a, since the washer models are noise free, even a very small value of epsilon results in high precision. This intuitively implies that the constraint of clusters being relatively equal in size holds true and the algorithm is able to detect corners precisely. However, the condition may be too strict to detect all corners. Thus, as slack variable $\varepsilon$ increases to 3, the recall for detection quickly rises to 80%. When the $\mathcal{K}$ nearest neighborhood is 20-30, a slack of $\varepsilon = 3$ reaches a maximum recall, however, for $\mathcal{K} = 15$ this value does not improve recall scores. Additionally, it is observed that for the entire parameter range, for $\mathcal{K} = 15$, recall value continues to be low. This can be logically explained as for $\mathcal{K} = 15$, if the number of clusters are $N = 3$, each cluster needs to have 5 points to be an exact corner. In this case, a slack of 3 points will result in a much larger bias in the cluster size with atleast one cluster left with only 2 points. This size of a cluster is not significant enough to detect a corner correctly. Thus, for a very small neighborhood, large values for $\varepsilon$ will result in poor performance. On the otherhand, for reasonably large neighborhoods of $\mathcal{K} = 20, 30$, varying $\varepsilon$ by a few points will drastically improve results.

In the next experiment, we vary $\theta_1$ and $\theta_2$ with $\varepsilon$ to show that different types of corners can have a variable variation in their mean curvature values. We vary $30^\circ < \theta_1 < 80^\circ$ and $110^\circ < \theta_2 < 140^\circ$ for $1 < \varepsilon < 6$ . All other parameters are kept fixed at $\mathcal{K} = 20$, and $\rho = 0.005$. The precision recall curve is shown in the graph in Fig. 6b which indicates that with a broader range for angular variation, the recall value immediately rises to 100%, ensuring that all types of corners are detected.

The corner detection algorithm is currently implemented in MATLAB R2016b and takes an average of 1.5 seconds for a point cloud of 27,000 points. We aim to increase this speed in future by implementing the algorithm in C++.

### C. *Automatic Weld Seam Detection*

In this section we introduce an application of the proposed algorithms for automation of robotic welding. Traditionally, welding seams (path that the robot follows during welding) for straight line joints are determined by importing an accurately drawn 3D CAD model of the workpiece into a CAD/CAM based software. The software extracts edges and the user is able to select the sequence for the robot to weld the joint. We present a novel approach that is based on detecting edges and corners directly from a point cloud that can be used to generate weld seams, eliminating the requirement of accurate 3D drawing and offline programming.

We test our method on 3D pointcloud of panel workpiece, captured using Asus Xtion. The panel workpiece only consist of straight line seams. For experimental evaluation, the point clouds have been cropped at a fixed size of 0.5x0.5x0.5m, to remove the surrounding walls and floor. We label the corners manually in Meshlab for groundtruth. Additionally, we compare our corner detection results with Harris 3D as a baseline, available from PCL.

For the proposed corner detector, $\rho$ was fixed at 0.005m, $\theta_1 = 60^\circ, \theta_2 = 140^\circ$ and $1 < \varepsilon < 6$ for $\mathcal{K} = 20$ nearest neighbors. For Harris 3D, we found the optimal threshold to be 0.01 while the radius for search was varied from $0.01 < r < 0.03$. We observed that highest precision for Harris 3D was achieved at $r = 0.01$ but with in a very small parameter shift for $r = 0.03$, the precision drops as a high number of FPs are generated. However, recall grows as the algorithm is able to detect almost all corners. As opposed to this, the precision for our proposed algorithm continues to stay high for the entire range of $\varepsilon$ indicating that very few FP corners were generated. Additionally, with a smaller value of $\varepsilon$, only a few corners were detected, however, as $\varepsilon$ is increased, all corners from the workpiece are successfully detected. The precision and recall curve for these parameters is shown in Fig. 7. It can be concluded that our proposed approach can be used to select specific type of corners much more accuarately while the Harris 3D is more suitable for detecting all kinds of sharp features together. Visually, this is demonstrated in Fig. 8 where Harris 3D can detect corners but misses out a few desired corners for weld seam generation. On the other hand, our proposed corner detector can detect all desired corners from the panel point cloud.

## V. Conclusion

We present a novel edge and corner detection algorithm applicable for unorganized point clouds. We demonstrate comparative evaluations on RGB-D semantic segmentation and 3D washer models from ShapeNet dataset for edge and corner detection respectively. Both of our methods show high accuracy as compared to state-of-the-art algorithms. We also implemented the two algorithms on a 3D pointcloud of panel workpiece, demonstrating that weld seams can be generated without the traditional CAD models.

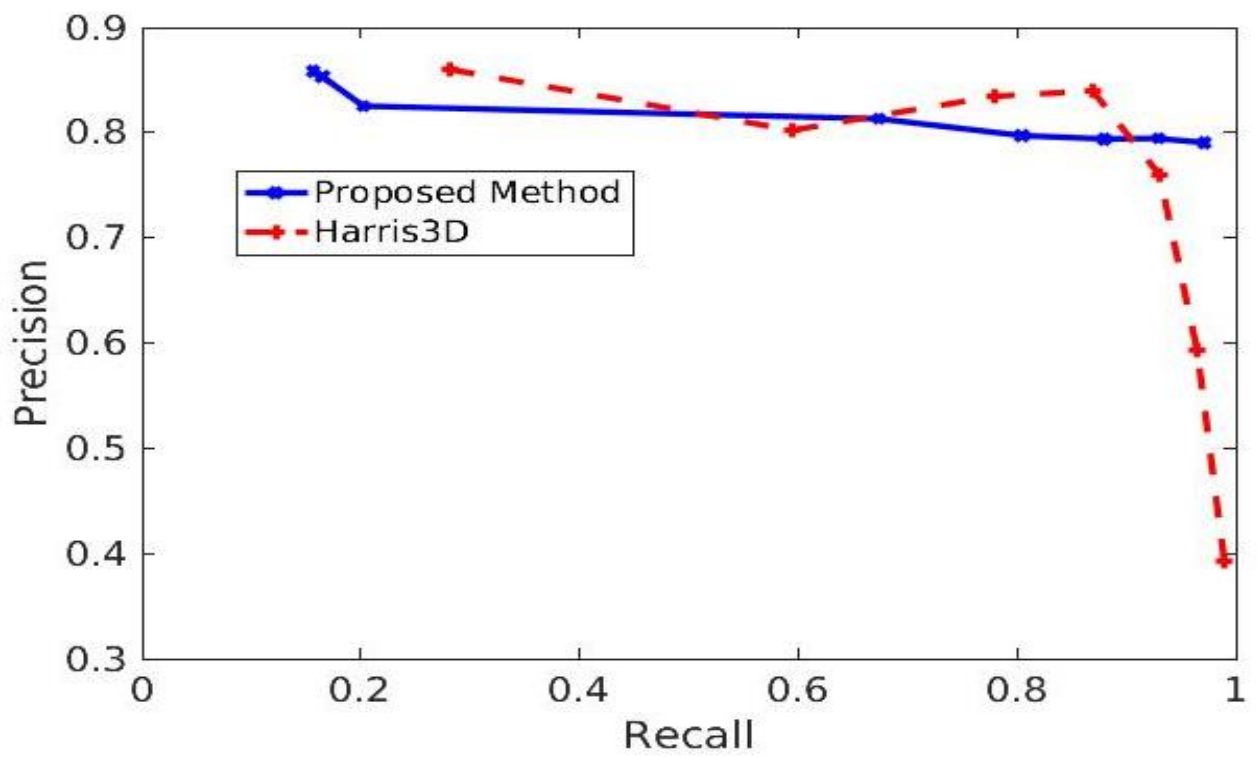

Fig. 7. Precision recall curve for Harris 3D and proposed corner detector on the panel dataset.

## ACKNOWLEDGMENT

The authors would like to thank the National Research Foundation, Keppel Corporation and National University of Singapore for supporting this research that is carried out in the Keppel-NUS Corporate Laboratory. The conclusions put forward reflect the views of the authors alone, and not necessarily those of the institutions within the Corporate Laboratory.

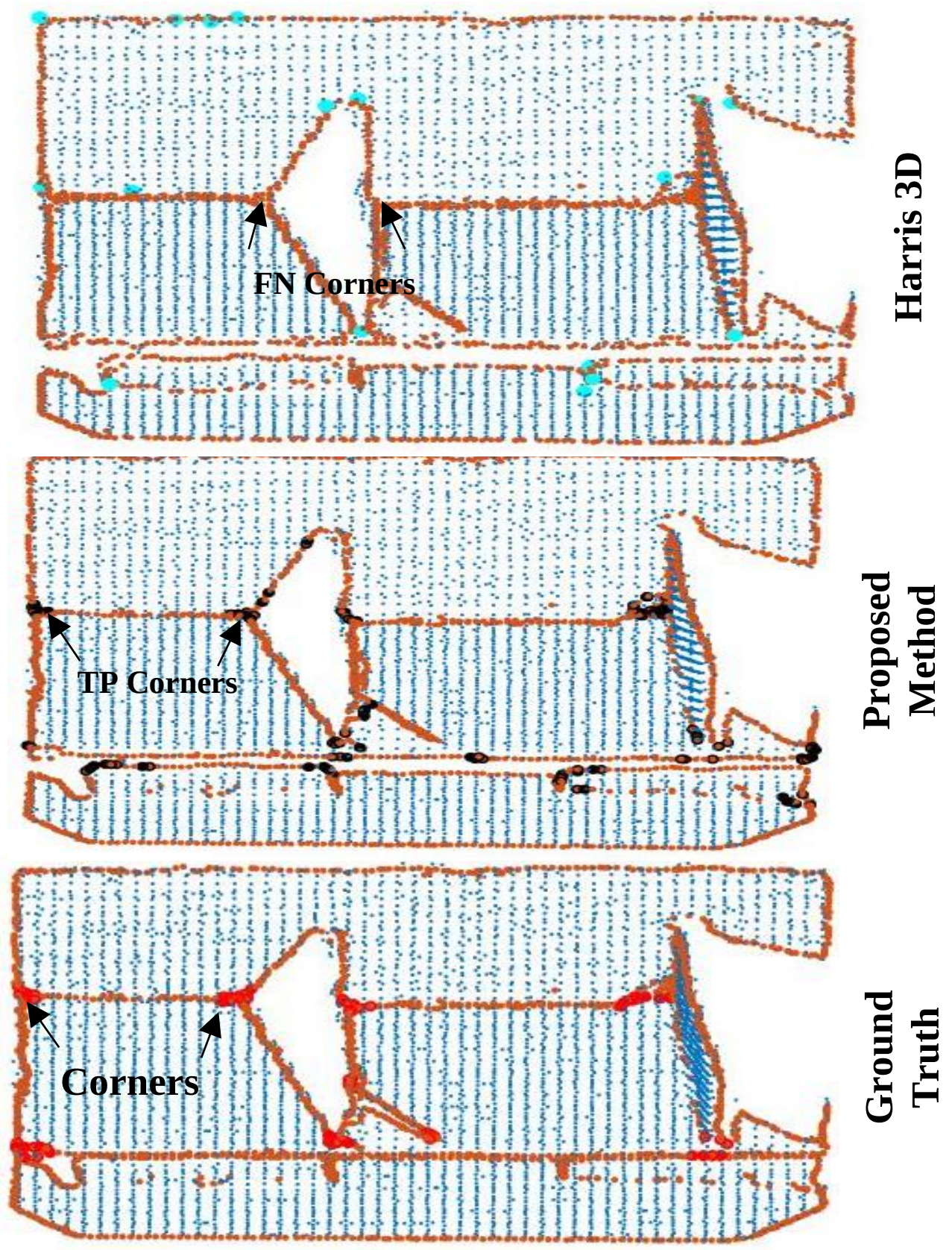

Fig. 8. Corner detection for panel workpiece using point clouds captured from Asus Xtion. a) results of corner detection using Harris 3D, b) results of corner detection from the proposed approach, c) manually labelled corners.